\documentclass[lettersize,journal]{IEEEtran}
\usepackage{amsmath,amsfonts}
\usepackage{algorithmic}
\usepackage{algorithm}
\usepackage{array}
\usepackage{subfigure}
\usepackage{multirow}
\usepackage[caption=false,font=normalsize,labelfont=sf,textfont=sf]{subfig}
\usepackage{textcomp}
\usepackage{stfloats}
\usepackage{url}
\usepackage{verbatim}
\usepackage{graphicx}
\usepackage{cite}
\usepackage{color}
\usepackage{CJKutf8}
\definecolor{ljcolor}{RGB}{34,139,34}

\usepackage{amsmath,amsfonts}
\usepackage{algorithmic}
\usepackage{array}
\usepackage[caption=false,font=normalsize,
   labelfont=sf,textfont=sf]{subfig}
\usepackage{textcomp}
\usepackage{stfloats}
\usepackage{url}
\usepackage{verbatim}
\usepackage{graphicx}
\usepackage{balance}
\begin{document}
\begin{CJK}{UTF8}{gbsn}

\title{TOE: A Grid-Tagging Discontinuous NER Model Enhanced by Embedding Tag/Word Relations and More Fine-Grained Tags}

\author{Jiang Liu, Donghong Ji, Jingye Li, Dongdong Xie, Chong Teng, Liang Zhao and Fei Li
\thanks{Jiang Liu, Donghong Ji, Jingye Li, Dongdong Xie, Chong Teng and Fei Li are with the Key Laboratory of Aerospace Information Security and Trusted Computing, Ministry of Education, School of Cyber Science and Engineering, Wuhan University (e-mail: liujiang@whu.edu.cn; dhji@whu.edu.cn; theodorelee@whu.edu.cn; 
xie.dongdong@whu.edu.cn; 
tengchong@whu.edu.cn;
foxlf823@gmail.com).
{\it{Corresponding author: Fei Li.}}
}
\thanks{Liang Zhao is with the University of São Paulo, Brazil (e-mail: zhao@usp.br).}
}

\markboth{Journal of \LaTeX\ Class Files,~Vol.~14, No.~8, August~2021}%
{Shell \MakeLowercase{\textit{et al.}}: A Sample Article Using IEEEtran.cls for IEEE Journals}


\maketitle



\begin{abstract}
So far, discontinuous named entity recognition (NER) has received increasing research attention and many related methods have surged such as hypergraph-based methods, span-based methods, and sequence-to-sequence (Seq2Seq) methods, etc.
However, these methods more or less suffer from some problems such as decoding ambiguity and efficiency, which limit their performance.
Recently, grid-tagging methods, which benefit from the flexible design of tagging systems and model architectures, have shown superiority to adapt for various information extraction tasks. 
In this paper, we follow the line of such methods and propose a competitive grid-tagging model for discontinuous NER.
We call our model TOE because we incorporate two kinds of Tag-Oriented Enhancement mechanisms into a state-of-the-art (SOTA) grid-tagging model that casts the NER problem into word-word relationship prediction.
First, we design a Tag Representation Embedding Module (TREM) to force our model to consider not only word-word relationships but also word-tag and tag-tag relationships.
Concretely, we construct tag representations and embed them into TREM, so that TREM can treat tag and word representations as queries/keys/values and utilize self-attention to model their relationships.
On the other hand, motivated by the Next-Neighboring-Word (NNW) and Tail-Head-Word (THW) tags in the SOTA model, we add two new symmetric tags, namely Previous-Neighboring-Word (PNW) and Head-Tail-Word (HTW), to model more fine-grained word-word relationships and alleviate error propagation from tag prediction.
In the experiments of three benchmark datasets, namely CADEC, ShARe13 and ShARe14, our TOE model pushes the SOTA results by about 0.83\%, 0.05\% and 0.66\% in F1, demonstrating its effectiveness.

\end{abstract}


\section{Introduction}
\begin{figure}[t]
\centering
\includegraphics[scale=0.3]{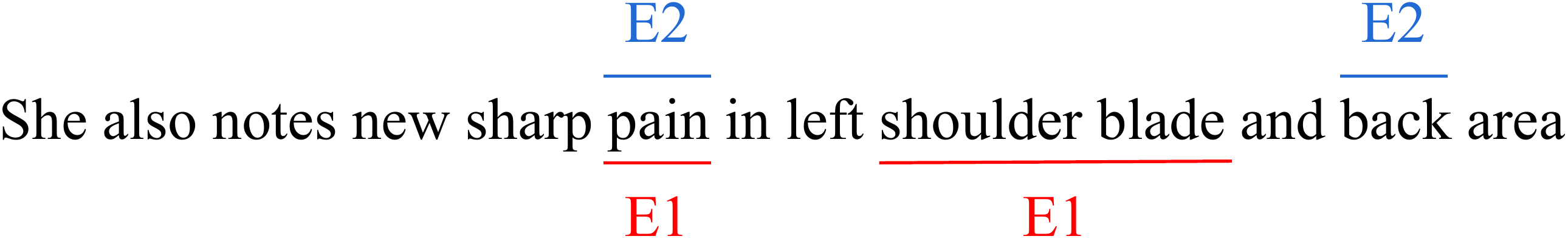}
\caption{An example to show two discontinuous entities in a sentence of the clinical corpus \cite{ref15}.
}
\label{fig1}
\end{figure}

\IEEEPARstart{N}amed entity recognition (NER) is a fundamental task for natural language processing (NLP),
which is able to facilitate many other NLP tasks (e.g., question answering \cite{ref1}, entity relationship extraction \cite{ref2}).
NER has been extensively studied and researchers have come up with numerous effective methods \cite{ref3, ref4, ref5,ref6,ref7}. 
Previously, most methods \cite{ref8,ref9,ref10,ref11,ref12,ref13} treat it as a sequential marking problem, in which each token is assigned with a tag representing its entity type. 
Their basic assumption is that entity mentions should be short text spans \cite{ref14} and should not overlap with each other. 
Although this assumption is valid in most cases, it is not always true, especially in clinical corpora. \cite{ref15}.
As shown in Figure \ref{fig1}, the two entities consist of several discontinuous segments and some segments are overlapped.
Therefore, it is necessary to design models that can recognize both flat entities and discontinuous entities.

To achieve this goal, some recent studies have developed some models for discontinuous NER, which can be roughly divided into the following categories:
\textbf{1)} Sequence-tagging-based methods \cite{ref16} extend the BIO tag scheme to more complex tag schemes such as BIOHD, but such ad hoc design is not flexible enough to handle all the situations.
\textbf{2)} Hypergraph-based methods represent all entity segments as graph nodes and learn to combine these nodes with individual classifiers, but such methods suffer from the false structure and structural ambiguity in the prediction process. 
\textbf{3)} Seq2Seq-based methods \cite{ref17, ref18} generate various entities directly, which unfortunately may suffer from decoding efficiency issues and certain common pitfalls of the Seq2Seq architecture, such as exposure bias.
\textbf{4)} Span-based methods \cite{ref19} list all possible spans and classify them according to the level of spans. However, these methods are limited by the maximum span length and result in considerable computational complexity due to span enumeration. 

Recently, grid-tagging-based methods achieve promising performance for discontinuous NER.
Wang et al. (2021) \cite{ref20} predict the entity boundaries and entity word relationships respectively through two grids and then decode the whole entities from the entity segment graph through maximal clique discovery. The latest state-of-the-art (SOTA) method, proposed by Li et al. 2022 \cite{ref21}, is also based on grid tagging. 
It transforms discontinuous NER into the word-word relationship recognition problem, and utilizes one grid to include all word-word relationships.

\begin{figure}[t]
\centering
\includegraphics[scale=0.4]{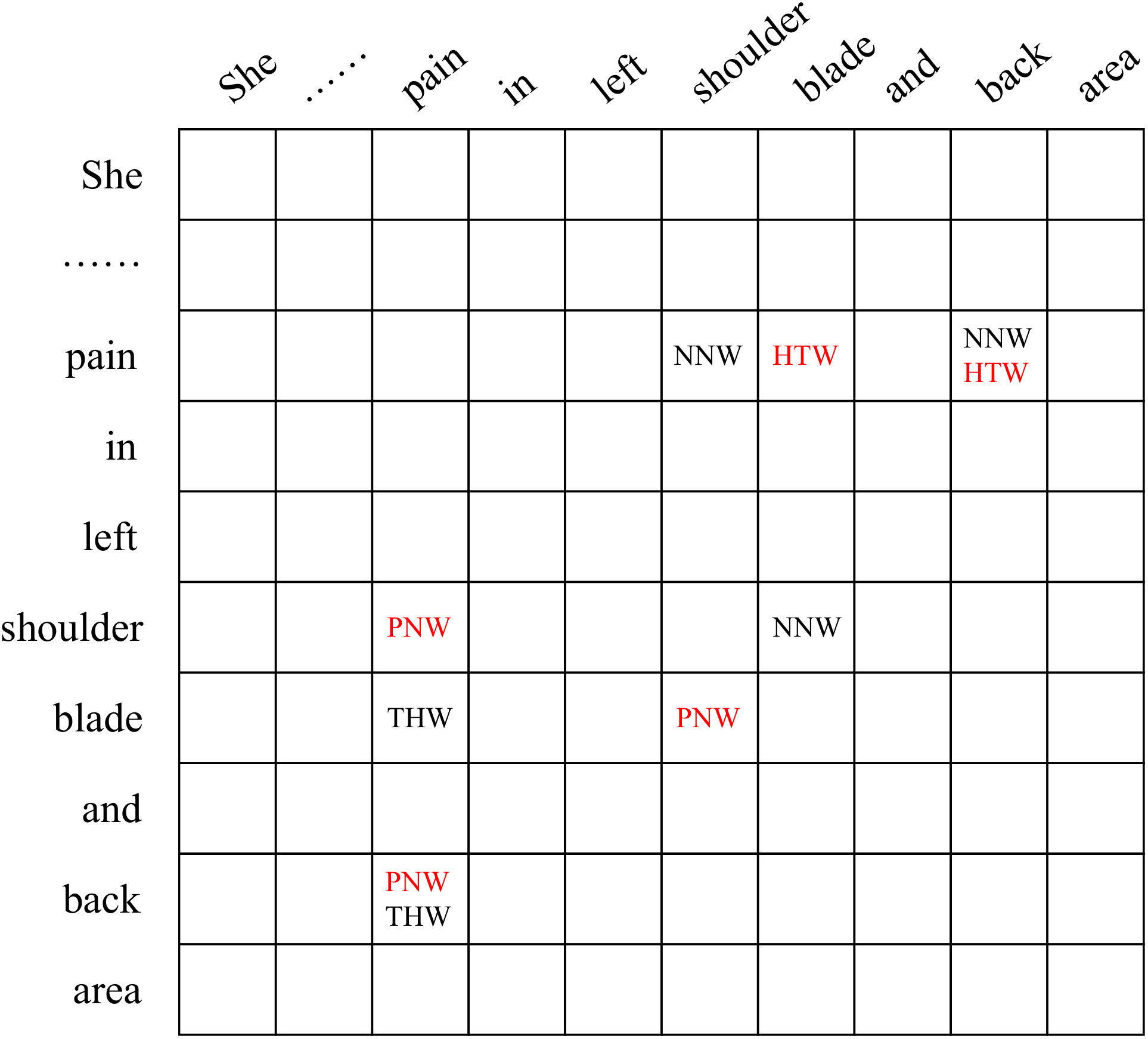}
\caption{An example to show our extended tag system. 
The red tags are new and the black ones are used in the SOTA model \cite{ref21}. Columns and rows represent the subjects and objects of the relationships, where a subject indicates the tail of the relationship arc and an object indicates the head. For more details refer to Section \ref{sec:tag_system}.
}
\label{fig2}
\end{figure}

In this paper, we follow the line of grid tagging for discontinuous NER and propose two tag-oriented enhancements to optimize the SOTA model. 
First, since the SOTA model only pays attention to the relationships between words, we embed tag representations into the model in order to model the relationships between each pair of tags and the relationships between tags and words.
These relationships are also important because entity mentions can be correctly identified only if all the tags have been predicted.
Second, we add two tags, namely {\tt{Previous-Neighboring-Word}} (\textbf{PNW}) and {\tt{Head-Tail-Word}} (\textbf{HTW}), to its tag system (including only two tags, {\tt{Next-Neighboring-Word}} (\textbf{NNW}) and {\tt{Tail-Head-Word}} (\textbf{THW})), in order to model more fine-grained word-word relationships.
As shown in Figure \ref{fig2}, when the model fails to recognize the HTW relationship between “pain” and “blade”, their THW relationship can still be recognized to compensate for such error.

\begin{figure*}[t]
\centering
\includegraphics[scale=0.55]{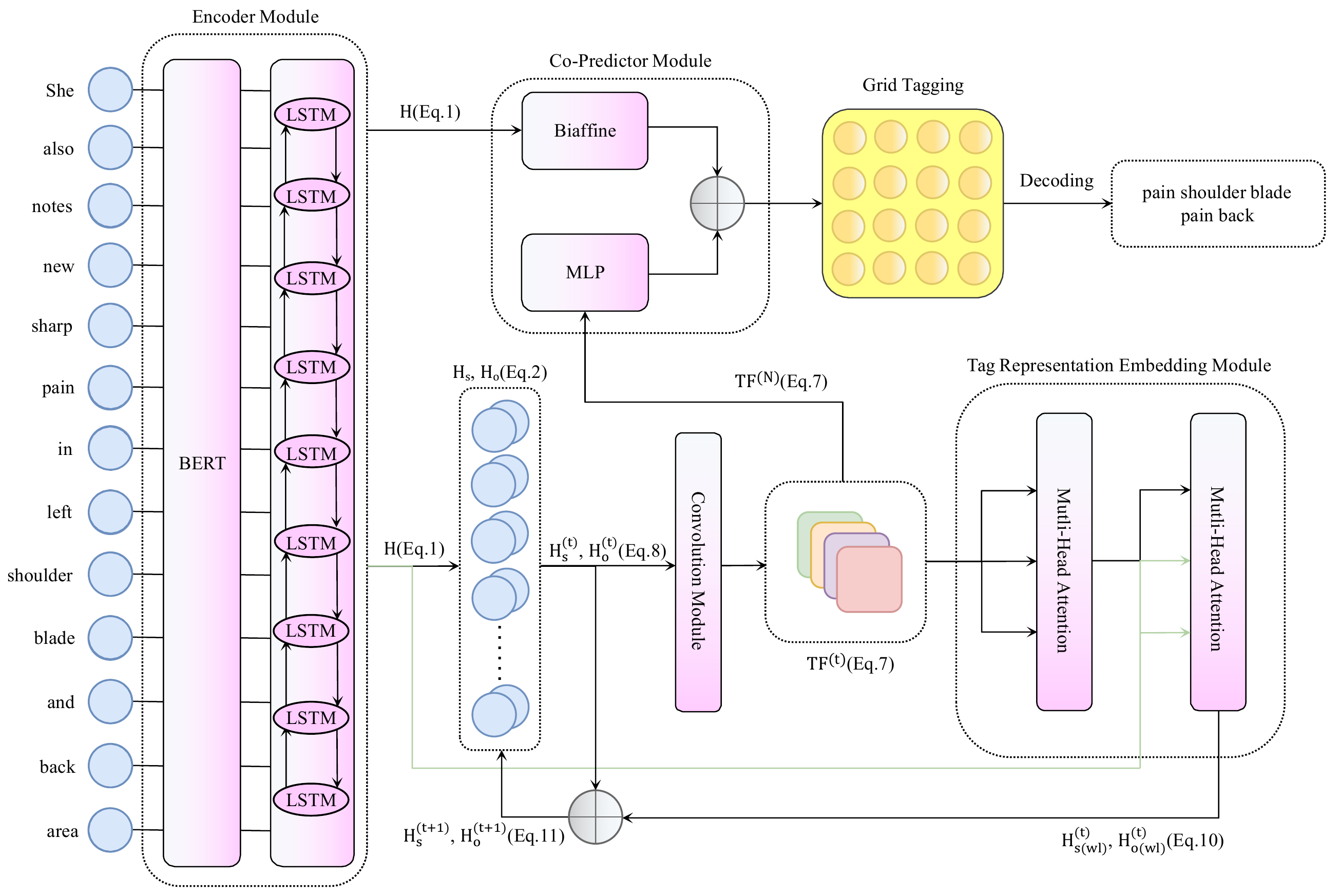}
\caption{The overall architecture of our TOE model. 
$H$ denotes the word representations.
$TF^{(t)}$ denotes the tag-aware grid features, where $t$ means that the tag representation embedding process may run several iterations. 
$ \bigoplus $ represents the element-level summation. 
} 
\label{fig3} 
\end{figure*}

To implement our above idea, we build a grid-tagging model with Tag-Oriented Enhancement (\textbf{TOE}).
The main framework of the model is shown in Figure \ref{fig3}. First, word representations are generated by BERT \cite{ref22} and BiLSTM \cite{ref23}. 
Then, we construct a tag representation embedding module (\textbf{TREM}) to embed tag representations into our model.
TREM employs a convolution layer to represent each tag as a two-dimensional (2D) table, because the tag in our task denotes the relationship of each word pair.
Then TREM uses self-attention to model the relationships between each pair of tags and the relationships between tags and words.
TREM can run multiple times to iteratively mix tag representations with word representations.
Finally, a co-predictor predicts the four word-word relationships that we defined in this paper, namely PNW, HTW, NNW and THW, using the word and tag representations jointly.
After that, discontinuous entities can be decoded out from these tags.

We conduct experiments on three datasets, namely CADEC \cite{ref5}, ShARe13 \cite{ref2} and ShARe14 \cite{ref24}.
Results show that our model achieves the best performance on all datasets and outperforms eight baselines including the SOTA model.
Our contributions can be summarized as:

• We propose a novel tag representation embedding module (TREM) to inject tag features and model the relationships across tags and words.

• We extend the tag system of the SOTA model to model more fine-grained word-word relationships and reduce error propagation.

• Our model achieves the SOTA performance on 3 benchmark datasets. We conduct substantial experiments on 3 datasets to analyze and understand our model.\footnote{The code is publicly available at https://github.com/solkx/TOE.git}


\section{Related Work}

We summarize the related methods for discontinuous NER in the following categories.
\noindent {\bf{Sequence-tagging-based methods}} In the field of NLP, NER is usually considered as a sequence tagging problem \cite{ref25,ref26,ref27}. Based on well-designed features, CRF based models have achieved leading performance \cite{ref28,ref29,ref30}. Recently, neural network models have been used for feature representation \cite{ref31,ref32}. In addition, upper and lower cultural lexical representations such as ELMo \cite{ref33}, Flair \cite{ref34} and BERT \cite{ref22} have also achieved great success. For NER, the end-to-end bidirectional LSTM-CRF model \cite{ref23,ref35,ref36} is a representative architecture. These models can only recognize regularly named entities. Tang et al. (2018) \cite{ref16} extended the BIO tagging scheme to BIOHD to solve the problem of discontinuous mention. Even so, there is still the problem of decoding ambiguity.

\noindent {\bf{Hypergraph-based methods}} Lu and Roth (2015) \cite{ref37} first proposed a model based on hypergraph method to solve NER, and expressed possible references by exponential method. Subsequent studies \cite{ref14,ref38,ref39} also developed and improved the method. For example, Muis and Lu (2018) \cite{ref14} used this method to deal with discontinuous NER, and Wang and Lu (2018) \cite{ref39} used a deep neural network to strengthen the hypergraph model.

\noindent {\bf{Seq2Seq-based methods}} Gillick et al. (2015) \cite{ref40} was the first to use the Seq2Seq model to solve NER. The model takes the original sentence as the input and takes the head and tail position, span length and entity type of all entities as the output. Fei et al. (2021) \cite{ref18} combined Seq2Seq and pointer network to deal with discontinuous NER. A recent study \cite{ref17} deals with all types of NER through the Seq2Seq model of pointer network based on BART \cite{ref41}, and generates the index and type sequence from the beginning to the end of all possible entities. However, the Seq2Seq model has potential decoding efficiency problems and exposure bias problems.

\noindent {\bf{Span-based methods}} Other studies deal with NER by identifying entity spans, that is, enumerating all possible entity spans, removing invalid entity spans or entity types, and finally retaining the final prediction results \cite{ref42,ref43}. Li et al. (2020a) \cite{ref44} redefine NER as a machine reading comprehension (MRC) task, ask questions for different entity types and extract entities according to the corresponding answers. Li et al. (2021a) \cite{ref19} convert the discontinuous NER to find the complete subgraph from the span-based entity segment graph, and obtain the competitive result. Unfortunately, due to enumeration, the effect of these methods is affected by the maximum span length and has considerable complexity, especially for longer entities.

\noindent {\bf{Grid-tagging-based methods}} Recently, the method based on grid marking \cite{ref20,ref21} has had a good performance. It transforms sentences into 2D tables. The method in Wang et al. (2021) \cite{ref20} includes three steps: 1) identifying the span of entity segments by marking the head and tail words in the table; 2) Extracting the relationship between entity segment span pairs by marking the head and tail words in another table; 3) The integrated entities are decoded from the entity segment span graph through maximum clique discovery. In contrast, the most advanced method \cite{ref21} uses a simpler process (two relationships, one table, no block decoding) and is more end-to-end, reducing error propagation. It obtains the features between words through the convolution layer, tags them on the grid, and identifies all possible entities through neighbor words relationships and head-tail relationships. The biggest difference between this method and other previous methods is that it focuses on the relationship between words rather than more accurate entity boundary recognition. In addition, the grid marking method can better avoid the disadvantages of some other methods, such as the disadvantages of sequence-to-sequence method and span based method.

\noindent {\bf{The differences between our model and previous models}}
Our model follows the SOTA model \cite{ref21} for discontinuous NER, which is also based on grid tagging. However, the differences include:
1) We design a new module to embed tag representations into our model to enhance the interactions between tags and words.
2) We extend the tag system in \cite{ref21} with two additional tags to model more fine-grained word-word relationships.


\section{Methodology}
We define the discontinuous NER task as a grid tagging problem and identify all possible entities through four predefined tags. 
Our model architecture is shown in Figure \ref{fig3}. It is mainly composed of four components and a tag system. The four components are the encoder module, the convolution module, the tag representation embedding module and the co-predictor module. Firstly, the encoder module is composed of a pre-training language model BERT \cite{ref22} and a bidirectional LSTM \cite{ref23}, which is used to generate the word representation of upper and lower culture from the input sentence. Then, the representation of multiple words on the grid is established and refined through the convolution module. Then, the tag features are captured by self-attention mechanism in TREM. The convolution module and TREM undergo multiple iterations to obtain more detailed features. Then, the co-predictor module \cite{ref45} is used to jointly infer the relationship between all word pairs. Finally, all possible entities are obtained by decoding.

\subsection{Encoder Module}
We use BERT \cite{ref22} as the text encoder of our model. Give an input sentence $ X=\{x_{1},x_{2},\ldots,x_{N}\} $, and input them into a pre-trained BERT. The BERT encoded by the multi-layer self-attention structure outputs the context representation of each context tag. To further enhance context modeling, we adopted bidirectional LSTM \cite{ref23} based on previous work \cite{ref19,ref46}. After BERT encoding, the sentence $X$ can be represented as: 
\begin{equation}
\label{deqn_ex1a}
\boldsymbol{H}=\{\boldsymbol{h}_{1},\boldsymbol{h}_{2},\ldots,\boldsymbol{h}_{N}\},
\end{equation}
where $\boldsymbol{h}_{i}\in\mathbb{R}^{d_{h}}$ is the representation of the $i$-th word and $ d_{h} $ represents the dimension of a word representation.

\subsection{Convolution Module}
Since the relationships between words in this paper are directional, each word plays either a subject or object role in one relationship. The subject indicates the tail of the relationship arc and the object indicates the head. As shown in Figure \ref{fig2}, subjects and objects correspond to the elements in the column and row respectively. We transform word representations into the subject and object spaces as below:
\begin{equation}
\label{deqn_ex1a}
\begin{aligned}
\boldsymbol{H}_{s}=\boldsymbol{W}_{1}\boldsymbol{H}+\boldsymbol{b}_{1}=\{\boldsymbol{h}_{1}^{s},\boldsymbol{h}_{2}^{s},\ldots,\boldsymbol{h}_{N}^{s}\}, \\
\boldsymbol{H}_{o}=\boldsymbol{W}_{2}\boldsymbol{H}+\boldsymbol{b}_{2}=\{\boldsymbol{h}_{1}^{o},\boldsymbol{h}_{2}^{o},\ldots,\boldsymbol{h}_{N}^{o}\},
\end{aligned}
\end{equation}
where $\boldsymbol{h}_{i}^{s}$, $\boldsymbol{h}_{i}^{o}\in\mathbb{R}^{d_{h}}$ represent the subject and object representations of the $i$-th word, $\boldsymbol{W}_{1}$, $\boldsymbol{W}_{2}\in\mathbb{R}^{d_{h}\times d_{h}}$ and $\boldsymbol{b}_{1}$, $\boldsymbol{b}_{2}\in\mathbb{R}^{d_{h}}$ are trainable weights and biases respectively.

The convolution module is then used as a representation refiner.
Firstly, the Conditional Normalization Layer (CLN) \cite{ref47} is used to generate the representation of words on the grid, which can be regarded as a three-dimensional matrix $ \boldsymbol{V}\in\mathbb{R}^{N\times N\times d_{h}}$:, in which each element in the $V_{ij}$ grid represents a word pair $ (x_{i},x_{j}) $:
\begin{equation}
\label{deqn_ex1a}
\boldsymbol{V}_{ij} = CLN(\boldsymbol{h}_{i}^{s},\boldsymbol{h}_{j}^{o})=\gamma_{ij}\odot(\frac{\boldsymbol{h}_{j}^{o}-\mu}{\sigma})+\lambda_{ij},
\end{equation}
where $ \boldsymbol{h}_{i}^{s} $ is the condition of the normalized gain parameters $ \gamma_{ij}=\boldsymbol{W}_{\alpha}\boldsymbol{h}_{i}^{s}+\boldsymbol{b}_{\alpha} $ and $ \lambda_{ij}=\boldsymbol{W}_{\beta}\boldsymbol{h}_{i}^{s}+\boldsymbol{b}_{\beta} $. $\boldsymbol{W}_{\alpha}$, $\boldsymbol{W}_{\beta}\in\mathbb{R}^{d_{h}\times d_{h}}$ and $\boldsymbol{b}_{\alpha}$, $\boldsymbol{b}_{\beta}\in\mathbb{R}^{d_{h}}$ are trainable weights and biases respectively. $ \mu $ and $ \sigma $ are the mean and standard deviation across the elements of $\boldsymbol{h}_{j}^{o}$.

Then the grid representations are enriched by adding the relative word position information $ \boldsymbol{E}^{wp}\in\mathbb{R}^{N\times N\times d_{wp}} $ between each pair of words and the grid position information $ \boldsymbol{E}^{gp}\in\mathbb{R}^{N\times N\times d_{gp}} $ that distinguishes the upper and lower triangular areas, and then mix with the word pair information $ \boldsymbol{V}\in\mathbb{R}^{N\times N\times d_{h}} $ to obtain the position area perception representation $ \boldsymbol{C}\in\mathbb{R}^{N\times N\times d_{c}} $ through a multi-layer perceptron (MLP):
\begin{equation}
\label{deqn_ex1a}
\boldsymbol{C} = MLP_{1}([\boldsymbol{V};\boldsymbol{E}^{wp};\boldsymbol{E}^{gp}]).
\end{equation}

Afterwards, 
the multiple 2D dilated convolutions (DConv) with different dilation rates are used to capture the interactions between the words with different distances, formulated as:
\begin{equation}
\label{deqn_ex1a}
\boldsymbol{Q} = GeLU(DConv(\boldsymbol{C})),
\end{equation}
where $ \boldsymbol{Q}\in\mathbb{R}^{N\times N\times d_{q}} $ is the output and $GeLU$ is a activation function \cite{ref48}.

\subsection{Tag Representation Embedding Module (TREM)}

The TREM module is used to
embed the tag representations into our model in order to model the interactions between tags as well as tags and words:
First, we generate the tag-aware grid feature $ \boldsymbol{TF}_{l}\in\mathbb{R}^{N\times N\times d_{t}} $ by mapping the grid representation $ \boldsymbol{Q} $ into the tag space.
Specifically, for the element $(i,j)$ in the grid corresponding to the word pair $ (x_{i}, x_{j}) $, we generate its tag-aware feature as:
\begin{equation}
\label{eq:q2tf}
\boldsymbol{TF}_{l}(i,j)=\boldsymbol{W}_{l} \boldsymbol{Q}_{ij} + \boldsymbol{b}_{l},
\end{equation}
where $\boldsymbol{W}_{l}\in\mathbb{R}^{d_{t}\times d_{q}}$ and $\boldsymbol{b}_{l}\in\mathbb{R}^{d_{t}}$ are trainable weights and biases.

Since there are four kinds of tags in this paper, namely NNW, PNW, HTW and THW (cf. Section \ref{sec:tag_system}), 
we concatenate them together as below: 
\begin{equation}
\label{eq:tf_concat}
\boldsymbol{TF}^{(t)}=[\boldsymbol{TF}_{NNW}^{(t)};\boldsymbol{TF}_{PNW}^{(t)};\boldsymbol{TF}_{HTW}^{(t)};\boldsymbol{TF}_{THW}^{(t)}],
\end{equation}
where $t$ means that the TREM module may run several times to refine $\boldsymbol{TF}\in \mathbb{R}^{N\times N\times 4d_{t}}$.
Theoretically, the number $M_{num}$ of tag space mappings can be smaller or larger than the number of tags, because our formulations in Equation \ref{eq:q2tf} and \ref{eq:tf_concat} are not constrained by this number.
However, we set $M_{num}$ the same as the number of tags heuristically since we consider $\boldsymbol{TF}_{l}$ as a tag representation.
We will empirically show the rationality of such method in the experiments (cf. Table \ref{tab:table4}).


We input $ \boldsymbol{TF}^{(t)} $ into the max-pooling layers ($Maxpool_{1}$, $Maxpool_{2}\in\mathbb{R}^{N\times 4d_{t}}$) and FFN layers to
recover the subject and object word features $ \boldsymbol{H}_{s}^{(t)} $ and $ \boldsymbol{H}_{o}^{(t)} $ at the $t$-th iteration:
\begin{equation}
\label{deqn_ex1a}
\begin{aligned}
\boldsymbol{H}_{s}^{(t)}=Maxpool_{1}(\boldsymbol{TF}^{(t)})\boldsymbol{W}_{s}+\boldsymbol{b}_{s},\\
\boldsymbol{H}_{o}^{(t)}=Maxpool_{2}(\boldsymbol{TF}^{(t)})\boldsymbol{W}_{o}+\boldsymbol{b}_{o}.
\end{aligned}
\end{equation}
where $\boldsymbol{W}_{s}$, $\boldsymbol{W}_{o}\in\mathbb{R}^{4d_{t}\times d_{h}}$ and $\boldsymbol{b}_{s}$, $\boldsymbol{b}_{o}\in\mathbb{R}^{d_{h}}$ are trainable weights and biases. $Maxpool_{1}$ and $Maxpool_{2}$ merge the representations $\boldsymbol{TF}^{(t)}$ along the rows and columns of the table respectively, so as to restore the subject and object word representations, $\boldsymbol{H}_{s}^{(t)}$ and $\boldsymbol{H}_{o}^{(t)}$.

Then we use the multi-head self-attention \cite{ref49} to mine the relationships between these tag-aware word representations:
\begin{equation}
\label{deqn_ex1a}
\begin{aligned}
\boldsymbol{H}_{s(ll)}^{(t)}= MultiHeadAttention(\boldsymbol{H}_{s}^{(t)},\boldsymbol{H}_{s}^{(t)},\boldsymbol{H}_{s}^{(t)}),\\
\boldsymbol{H}_{o(ll)}^{(t)}= MultiHeadAttention(\boldsymbol{H}_{o}^{(t)},\boldsymbol{H}_{o}^{(t)},\boldsymbol{H}_{o}^{(t)}),
\end{aligned}
\end{equation}
and another multi-head self-attention to mine the relationships between the original word representations and these tag-aware word representations:
\begin{equation}
\label{deqn_ex1a}
\begin{aligned}
\boldsymbol{H}_{s(wl)}^{(t)}= MultiHeadAttention(\boldsymbol{H}_{s(ll)}^{(t)},\boldsymbol{H}_{s},\boldsymbol{H}_{s}),\\
\boldsymbol{H}_{o(wl)}^{(t)}= MultiHeadAttention(\boldsymbol{H}_{o(ll)}^{(t)},\boldsymbol{H}_{o},\boldsymbol{H}_{o}).
\end{aligned}
\end{equation}



Since the TREM module may run several times to iteratively refine the tag-aware representations,
we add a residual connection \cite{ref50} to alleviate the gradient vanishment problem:
\begin{equation}
\label{deqn_ex1a}
\begin{aligned}
\boldsymbol{H}_{s}^{(t+1)}= LayerNorm(\boldsymbol{H}_{s}^{(t)}+\boldsymbol{H}_{s(wl)}^{(t)}),\\
\boldsymbol{H}_{o}^{(t+1)}= LayerNorm(\boldsymbol{H}_{o}^{(t)}+\boldsymbol{H}_{o(wl)}^{(t)}),
\end{aligned}
\end{equation}
where these new features are fed back to the convolution module for next iteration.


\subsection{Co-Predictor Module}
After the TREM, we get the tag-aware grid features $ \boldsymbol{TF}^{(N)} $ for each word pair. These features are fed into an MLP to predict the relationships between each pair of words. 
In addition, we enhance the relational classification by combining the MLP predictor with a biaffine predictor. 
Therefore, we take these two predictors to calculate the two independent relationship distributions $ (x_{i},x_{j}) $ of word pairs at the same time, and combine them as the final prediction. For MLP, its input is the output $ \boldsymbol{TF}^{(N)} $ of TREM, so the relationship score of each word pair $ (x_{i},x_{j}) $ is calculated as:
\begin{equation}
\label{deqn_ex1a}
\boldsymbol{y}^{'}_{ij}= MLP_{2}(\boldsymbol{TF}^{(N)}(i,j)),
\end{equation}

The input of the biaffine predictor is the output $ \boldsymbol{H} $ of the encoder layer, which can be considered as a residual connection \cite{ref50}. Two MLPs are used to calculate the representation of each word in the word pair $ (x_{i},x_{j}) $. Then, the relationship score between word pairs $ (x_{i},x_{j}) $ is calculated using a biaffine classifier \cite{ref51}:
\begin{equation}
\label{deqn_ex1a}
\boldsymbol{y}^{''}_{ij}=\boldsymbol{s}_{i}^{\top}\boldsymbol{U}\boldsymbol{o}_{j}+\boldsymbol{W}[\boldsymbol{s}_{i};\boldsymbol{o}_{j}]+\boldsymbol{b},
\end{equation}
where $ \boldsymbol{U} $, $ \boldsymbol{W} $ and $ \boldsymbol{b} $ are trainable parameters, and $ \boldsymbol{s}_{i}=MLP_{3}(\boldsymbol{h}_{i}^s) $ and $ \boldsymbol{o}_{j}=MLP_{4}(\boldsymbol{h}_{j}^o) $ represent the subject and object representations respectively. 
Finally, we combine the scores from the MLP and biaffine predictors to get the final score:
\begin{equation}
\label{deqn_ex1a}
\boldsymbol{y}_{ij}=Softmax(\boldsymbol{y}_{ij}^{'}+\boldsymbol{y}_{ij}^{''}).
\end{equation}

\begin{algorithm}[!t]
\caption{Decoding Algorithm.}\label{alg:alg1}
\begin{algorithmic}
\STATE \textbf{Input:} The relationships $R\in\mathbb{R}^{N\times N\times l_{n}}$ of all the word pairs, where $l_{n}$ is the number of word relationship tags. $R_{ij}^{l}$ indicates that the word pair ($x_{i}$, $x_{j}$) has an $l$ relationship, where $i, j \in [1, N]$.
\STATE \textbf{Output:} Entity set E.
\STATE 1: $E=[]$
\STATE 2: \textbf{for} $R_{ij}^{l} \in R$ and $ j \geq i$ \textbf{do}
\STATE 3: \hspace{0.5cm}\textbf{if} $Exist(R_{ji}^{THW})$ \textbf{or} $Exist(R_{ij}^{HTW})$ \textbf{then}
\STATE 4: \hspace{1cm}$S=[i]$   // Store the head word index
\STATE 5: \hspace{1cm}\textbf{if} $i=j$  \textbf{then} // Entity contains only one word
\STATE 6: \hspace{1.5cm}$E.add(S)$ // Store the entity span to $E$
\STATE 7: \hspace{1cm}\textbf{else} // Find next entity word
\STATE 8: \hspace{1.5cm}\textbf{for} $k\in (i, j]$ \textbf{do}
\STATE 9: \hspace{2cm}$FindNext(S, R_{ik}^{NNW}, R_{ki}^{PNW}, k, j, E)$
\STATE 10: \textbf{return} $E$
\STATE 11: // Find next entity word $m$ based on $r_{1}$ and $r_{1}$
\STATE 12: \textbf{function} $FindNext(S, r_{1}, r_{2}, m, e, E)$
\STATE 13: \hspace{0.5cm}\textbf{if} $Exist(r_{1})$ and $Exist(r_{2})$ \textbf{then}
\STATE 14: \hspace{1cm}$S.add(m)$ // Add the next word index
\STATE 15: \hspace{0.5cm}\textbf{if} $m=e$ \textbf{then} // Next word is the tail word
\STATE 16: \hspace{1cm}$E.add(S)$ // Store the entity span to $E$
\STATE 17: \hspace{0.5cm}\textbf{else} // Recursively find next entity word
\STATE 18: \hspace{1cm}\textbf{for} $k\in(m,e]$ \textbf{do}
\STATE 19: \hspace{1.5cm}$FindNext(S, R_{mk}^{NNW}, R_{km}^{PNW}, k, e, E)$
\end{algorithmic}
\label{alg1}
\end{algorithm}

\begin{figure}[t]
\centering
\includegraphics[scale=0.6]{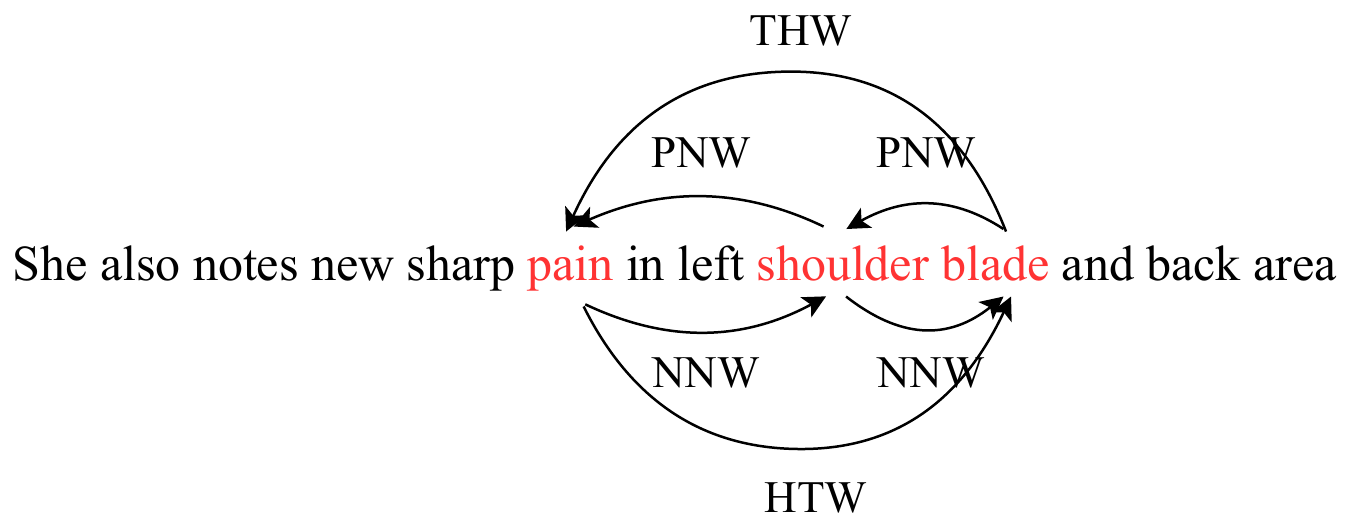}
\caption{An example to show the process of recognizing ``pain shoulder blade".
} 
\label{fig6} 
\end{figure}

\subsection{Our Tagging System}
\label{sec:tag_system}
In the SOTA model \cite{ref21}, two kinds of tags are predicted: 

• {\tt{Next-Neighboring-Word}} (NNW) indicates that the word pair $ (x_{i},x_{j}) $ belongs to an entity, and the next word of $ x_{i} $ in the entity is $ x_{j} $.

• {\tt{Tail-Head-Word}} (THW) indicates that the word in the row of the grid is the tail of the entity, and the word in the column of the grid is the head of the entity.


Although such tagging design is effective, it has some drawbacks. For example, when the model misses a THW relationship, it will fail to recognize the corresponding entity, which cannot be recovered.
Moreover, we believe that although their tagging design is elegant, it results in a sparse tag distribution in the grid and thus loses certain word-word relationships.
To enhance the tagging system and model more fine-grained word-word relationships, 
we propose two new tags:

• {\tt{Previous-Neighborhood-Word}} (PNW) indicates that the word pair $ (x_{i},x_{j}) $ belongs to an entity. The previous word of $ x_{i} $ in the entity is $ x_{j} $.

• {\tt{Head-Tail-Word}} (HTW) indicates that the word in the row of the grid is the head of the entity, and the word in the column of the grid is the tail of the entity.

By using these tags, we can model fine-grained word-word relationships and compensate certain error propagation from the model prediction.
For example, 
we jointly predict the NNW and PNW relationships, and 
when both of them exist, 
we think that the word pair belongs to the same entity.
Similarly, we jointly predict the THW and HTW relationships and when one of them exists, we think that the word pair is the head and tail of an entity. 
The advantage of using this decoding strategy will be shown in the ablation studies (cf. Table \ref{tab:table4}). 

Moreover, we show the pseudo-code of using this decoding strategy in Algorithm \ref{alg:alg1}. 
This decoding algorithm is mostly similar to the one used in Li et al. (2022) \cite{ref21}, 
while the differences exist in finding the head entity words (line 3) and non-head entity words (line 9). Because we add two new tags, PNW and HTW, the condition of head entity words changes from ``THW'' to ``HTW or THW'' and the condition of non-head entity words changes from ``NNW'' to ``NNW and PNW''.

Based on this decoding algorithm, we also give an example in Figure \ref{fig6} to explain the process of recognizing ``pain shoulder blade". By using the NNW relationship with the subject ``pain" and object ``shoulder" and the PNW relationship with the subject ``shoulder" and object ``pain", we recognize ``pain shoulder" as a part of the entity. 
Similarly, ``shoulder blade" is also recognized in the same way.
Then, by using the HTW and THW relationships, we recognize ``pain" and ``blade" are the head and tail of the entity, so that ``pain shoulder blade" can be recognized completely.



\subsection{Learning}
As shown in Figure \ref{fig2}, we can see that there may be more than one relationship between each pair of words.
Therefore, we adopt a cross-entropy loss function that is extended for multi-label classification \cite{ref52}. 
In order to predict the correct tag, we need the score of each target tag to be no less than that of each non-target tag. 
In addition, we define a threshold so that 
the scores of target classes are greater than the threshold, and the scores of non-target classes are less than the threshold. The final loss function is:
\begin{equation}
\label{deqn_ex1a}
\begin{aligned}
& \mathcal{L}=log \Big(1+\sum_{n,m}e^{\hat{y}_{(i,j)}^{n}-y_{(i,j)}^{m}}\\
& \qquad \qquad \qquad \qquad \ +\sum_{n}e^{\hat{y}_{(i,j)}^{n}-y_{0}}+\sum_{m}e^{y_{0}-y_{(i,j)}^{m}} \Big), \\
& \ =log \Big(e^{y_{0}}+\sum_{n}e^{\hat{y}_{(i,j)}^{n}} \Big)+log \Big(e^{-y_{0}}+\sum_{m}e^{-y_{(i,j)}^{m}} \Big),
\end{aligned}
\end{equation}
where $ n\in \Omega_{neg} $, $ m\in \Omega_{pos} $, and $ \Omega_{neg} $, $ \Omega_{pos} $ are the non-target and target tag sets respectively. $ \hat{y}_{(i,j)}^{n} $ and $ y_{(i,j)}^{m} $ are the non-target and target tag scores respectively. $y_{0}$ represents the threshold. This loss function is similar to the circle loss \cite{ref53}.
\begin{table*}[!t]
\caption{Statistics of three datasets. 
\label{tab:table1}}
\centering
\begin{tabular}{c c c c c| c c c c| c c c c}
\hline
{}&\multicolumn{4}{c}{{\bf{CADEC}}}&\multicolumn{4}{c}{{\bf{ShARe13}}}&\multicolumn{4}{c}{{\bf{ShARe14}}}\\
\cline{2-13}
{}&All&Train&Dev&Test&All&Train&Dev&Test&All&Train&Dev&Test\\
\hline
{\#Sentences}&7,597&	5,340&	1,097&	1,160&18,767&	8,508&	1,250&	9,009&34,614&	17,407&	1,361&	15,850\\
{\#Entities}&6,318&	4,430&	898&	990&11,148&	5,146&	669	&5,333&19,047&	10,354&	771&	7,922\\
{\#Discontinuous}&679&	491	&94&	94&1,088&	581	&71	&436&1,650&	1,004&	80&	566\\
{\%Discontinuous}&10.7&	11.1&	10.5&	9.5&9.8&	11.3&	10.6&	8.2&	8.7&9.7&	10.4&	7.1\\
\hline
\end{tabular}
\end{table*}

\begin{table}[!t]
\caption{Hyper-parameter settings \label{tab:table2}}
\centering
\begin{tabular}{c c}
\hline
Hyper-parameter & value\\
\hline \hline
$ d_{h} $ & 768\\
$ d_{wp} $ & 20\\
$ d_{gp} $ & 20\\
$ d_{q} $ & 64, 80, 96, 128\\
$y_{0}$&0\\
Dropout & 0.1, 0.3, 0.5\\
Learning rate (BERT) & 5e-6\\
Learning rate (others) & 1e-3\\
Batch size & 12, 16\\
Warm factor & 0, 0.1, 0.4\\
Rounds (TREM iterations) & 3  \\
\hline
\end{tabular}
\end{table}
\section{Experiment Setting}
\subsection{Datasets}
In order to evaluate our model, we conducted experiments on three discontinuous NER datasets, namely CADEC \cite{ref5}, ShARe13 \cite{ref2} and ShARe14 \cite{ref24}, all of which come from the documents in biomedical or clinical fields. They all contain only one entity type, in which the entity type in CADEC is {\tt{ADR}}, and the entity type in ShARe13 and ShARe14 is {\tt{Disease\_Disorder}}. We use the preprocessing script provided by Dai et al. (2020) \cite{ref54} for dataset segmentation. 
In these discontinuous NER datasets, discontinuous entities account for about 10\% of the total entities. The statistics of these datasets are shown in Table \ref{tab:table1}. 

\begin{table*}[!t]
\caption{Performance comparisons between the baseline models and our model on three datasets. The bold number represents the highest result in each column. 
We also perform the significance test on the F1s of our model and the SOTA model \cite{ref21} on the development set and test set.
$^*$ denotes significance at p$<$0.05. {The numbers in parentheses mean the model results on the development sets.}
\label{tab:table3}
}
\centering
\begin{tabular}{c c c c c c c c c c c}
\hline
\multicolumn{2}{c}{\multirow{2}{*}{{\bf{MODEL}}}} & \multicolumn{3}{c}{{\bf{CADEC}}} & \multicolumn{3}{c}{{\bf{ShARe13}}} & \multicolumn{3}{c}{{\bf{ShARe14}}}\\
\cline{3-11}
\multicolumn{2}{c}{} & {\bf{P}} & {\bf{R}} & {\bf{F1}} & {\bf{P}} & {\bf{R}} & {\bf{F1}} & {\bf{P}} & {\bf{R}} & {\bf{F1}}\\
\hline
{\bf{Sequence Tagging}}&Tang et al. (2018) \cite{ref16}&67.80&64.99&66.36&—&—&—&—&—&—\\
\hline
{\bf{Hypergraph-based}}&Wang and Lu (2019) \cite{ref55}&72.10 &	48.40 &	58.00 &	83.80 &	60.40 &	70.30 &	79.10 &	70.70 &	74.70\\
\hline
\multirow{2}{*}{{\bf{Seq2Seq}}}&Yan et al. (2021) \cite{ref17}&70.08 &	71.21 &	70.64 &	82.09 &	77.42 &	79.69 &	77.20 &	{\bf{83.75}} &	80.34 \\
\cline{2-11}
\multirow{2}{*}{}&Fei et al. (2021) \cite{ref18}&75.50 &	71.80 &	72.40 	&{\bf{87.90}} & 77.20 &	80.30 &	—&	—&—\\
\hline
{\bf{Span-based}}&Li et al. (2021a) \cite{ref19}&—&—&	69.90&—&—&	82.50 &—&	—&—\\
\hline
{\bf{Others}}&Dai et al. (2020) \cite{ref54}&68.90 &	69.00 &	69.00 &	80.50 &	75.00 &	77.70 &	78.10 &	81.20 &	79.60 \\
\hline
\multirow{3}{*}{{\bf{Grid Tagging}}}&Wang et al. (2021) \cite{ref20}&70.50 &	{\bf{72.50}} &	71.50 &	84.30 &	78.20 &	81.20 &	78.70 &	82.15 &	80.39 \\
\cline{2-11}
\multirow{3}{*}{}&\multirow{2}{*}{Li et al. (2022) \cite{ref21}}&74.09&72.35&73.21&85.57&79.68&82.52&79.88&83.71&81.75 \\
\multirow{3}{*}{}&\multirow{2}{*}{} &{(70.59)}&{({\bf{68.45}})}&{(69.50)}&{(80.20)}&{({\bf{77.83}})}&{(78.99)}&{(82.29)}&{(81.37)}&{(81.82)}\\
\hline
&\multirow{2}{*}{TOE (ours)}&{\bf{77.77}}&70.66&{\bf{74.04}}$^*$&85.18&{\bf{80.12}}&{\bf{82.57}}&{\bf{82.26}}&82.57&{\bf{82.41}}$^*$\\
&\multirow{2}{*}{}&{({\bf{74.22}})}&{(67.79)}&{({\bf{70.86}}$^*$)}&{({\bf{81.38}})}&{(77.59)}&{({\bf{79.42}}$^*$)}&{({\bf{83.77}})}&{({\bf{82.22}})}&{({\bf{82.98}}$^*$)}\\
\hline
\end{tabular}
\end{table*}

\subsection{Baselines}
{\bf{Sequence-tagging-based methods}} assign a tag to each token with different tag schemes, such as BIOHD \cite{ref16}. {\bf{Span-based methods}} enumerate all possible spans and combines them into entities \cite{ref19}. {\bf{Hypergraph-based methods}} use hypergraphs to represent and infer entity mention \cite{ref55}. {\bf{Seq2Seq-based methods}} directly generate the word sequences of the entities at the decoder side \cite{ref17,ref18}. {\bf{Grid-tagging-based methods}} assign a tag for each pair of words and entities can be decoded out from these tags \cite{ref20,ref21}.
We also compare with other methods that cannot be grouped into the above categories, such as the transition-based method \cite{ref54}.

\subsection{Evaluation Metrics}
Our evaluation metrics follow previous work \cite{ref17,ref37,ref56}, using the precision (P), recall (R) and F1. If the token sequence and type of a predicted entity are exactly the same as those of a gold entity, the predicted entity is regarded as true-positive. We run each experiment three times and report their average value.

\subsection{Implementation Details}
Our hyper-parameter settings are given in Table \ref{tab:table2}. 
The hyper-parameters are adjusted according to the fine-tuning on the development sets.
In addition, since the datasets come from different fields, we use different pre-trained langauge models to generate word representations. 
For CADEC, we use BioBERT \cite{ref57}, and for ShARe13 and ShARe14, we use ClinicalBERT \cite{ref58}.
Moreover, we use AdamW \cite{ref59} as the optimizer. 
Our model is implemented using PyTorch and trained using NVIDIA RTX 3090 GPU. 

\begin{table*}[!t]
\caption{
Ablation experiments. 
We report the model performance when deleting some modules such as TREM and extending tags,
or changing some configurations such as the rounds of TREM, decoding methods and tag mapping number $M_{num}$. $T(l_{1},l_{2})$ indicates that we think the word relationship really exists when both $l_{1}$ and $l_{2}$ exist. $L(l_{1},l_{2})$ indicates that we think the word relationship really exists when either one of $l_{1}$ and $l_{2}$ exist. 
Best Setting: T(NNW,PNW),L(THW,HTW); Rounds = 3; $M_{num}$ = 4. {The numbers in parentheses mean the model results on the development sets.}
\label{tab:table4}
}
\centering
\begin{tabular}{l c c c c c c c c c}
\hline
{}&\multicolumn{3}{c}{{\bf{CADEC}}} & \multicolumn{3}{c}{{\bf{ShARe13}}} & \multicolumn{3}{c}{{\bf{ShARe14}}}\\
\cline{2-10}
{}&{\bf{P}} & {\bf{R}} & {\bf{F1}} & {\bf{P}} & {\bf{R}} & {\bf{F1}} & {\bf{P}} & {\bf{R}} & {\bf{F1}}\\
\cline{2-10}

\multirow{2}{*}{Best Setting}&77.77&70.66 &{\bf{74.04}}&85.18&80.12&{\bf{82.57}}&82.26&82.57&{\bf{82.41}}\\
\multirow{2}{*}{}&{({\bf{74.22}})}&{(67.79)}&{({\bf{70.86}})}&{(81.38)}&{(77.59)}&{({\bf{79.42}})}&{(83.77)}&{({\bf{82.22}})}&{({\bf{82.98}})}\\
\hline
\multirow{2}{*}{$w/o$ TREM}&{73.44}&{72.19}&{72.80}&{84.85}&{79.31}&{81.98}&{79.50}&{83.36}&{81.38}\\
\multirow{2}{*}{}&{(69.69)}&{({\bf{69.86}})}&{(69.77)}&{(79.67)}&{(77.44)}&{(78.53)}&{(82.24)}&{(79.90)}&{(81.06)}\\
\hline
\multirow{2}{*}{Rounds = 2}&{75.75}&{70.93}&{73.25}&{85.84}&{77.55}&{81.38}&{79.91}&{83.44}&{81.63}\\
\multirow{2}{*}{}&{(71.41)}&{(69.15)}&{(70.26)}&{(81.07)}&{(75.98)}&{(78.31)}&{(82.99)}&{(80.52)}&{(81.73)}\\
\cline{2-10}
\multirow{2}{*}{Rounds = 4}&{75.84}&{69.12}&{72.23}&{84.80}&{78.36}&{81.41}&{81.00}&{82.91}&{81.93}\\
\multirow{2}{*}{}&{(72.88)}&{(67.35)}&{(69.95)}&{(79.22)}&{(77.24)}&{(78.11)}&{(84.15)}&{(79.07)}&{(81.53)}\\
\hline
\multirow{2}{*}{NNW+THW}&{73.78}&{{\bf{72.62}}}&{73.20}&{84.33}&{79.68}&{81.94}&{77.72}&{{\bf{84.97}}}&{81.18}\\
\multirow{2}{*}{}&{(69.66)}&{(69.73)}&{(69.69)}&{(79.26)}&{({\bf{78.08}})}&{(78.67)}&{(81.82)}&{(82.19)}&{(81.99)}\\
\cline{2-10}
\multirow{2}{*}{PNW+HTW}&{74.91}&{71.47}&{73.12}&{84.89}&{79.46}&{82.07}&{78.81}&{84.27}&{81.44}\\
\multirow{2}{*}{}&{(70.54)}&{(68.50)}&{(69.49)}&{(80.20)}&{(77.83)}&{(78.99)}&{(82.29)}&{(81.37)}&{(81.82)}\\
\hline
\multirow{2}{*}{T(NNW,PNW),T(THW,HTW)}&{{\bf{78.10}}}&{68.37}&{72.90}&{{\bf{86.07}}}&{79.13}&{82.45}&{{\bf{83.02}}}&{81.72}&{82.37}\\
\multirow{2}{*}{}&{(73.35)}&{(66.33)}&{(69.65)}&{({\bf{81.47}})}&{(76.28)}&{(78.77)}&{({\bf{86.06}})}&{(77.58)}&{(81.60)}\\
\cline{2-10}
\multirow{2}{*}{L(NNW,PNW),T(THW,HTW)}&{77.93}&{67.38}&{72.23}&{85.01}&{79.52}&{82.17}&{82.58}&{81.24}&{81.91}\\
\multirow{2}{*}{}&{(73.33)}&{(66.07)}&{(69.48)}&{(80.56)}&{(77.04)}&{(78.73)}&{(84.66)}&{(78.11)}&{(81.25)}\\
\cline{2-10}
\multirow{2}{*}{L(NNW,PNW),L(THW,HTW)}&{76.18}&{70.66}&{73.28}&{84.55}&{{\bf{80.19}}}&{82.31}&{81.05}&{83.25}&{82.14}\\
\multirow{2}{*}{}&{(71.25)}&{(68.13)}&{(69.63)}&{(80.47)}&{(77.47)}&{(78.93)}&{(84.01)}&{(79.16)}&{(81.51)}\\
\hline
\multirow{2}{*}{$M_{num}$ = 2}&{75.58}&{69.87}&{72.61}&{85.27}&{77.88}&{81.38}&{80.85}&{81.39}&{81.11}\\
\multirow{2}{*}{}&{(71.80)}&{(68.28)}&{(70.00)}&{(80.39)}&{(76.18)}&{(78.20)}&{(84.49)}&{(79.90)}&{(82.13)}\\
\cline{2-10}
\multirow{2}{*}{$M_{num}$ = 6}&{76.26}&{70.28}&{73.14}&{84.54}&{79.09}&{81.72}&{79.83}&{83.36}&{81.55}\\
\multirow{2}{*}{}&{(71.88)}&{(67.57)}&{(69.65)}&{(79.41)}&{(76.78)}&{(78.07)}&{(83.33)}&{(80.12)}&{(81.69)}\\
\cline{2-10}
\multirow{2}{*}{$M_{num}$ = 8}&{76.04}&{69.19}&{72.44}&{85.29}&{78.26}&{81.62}&{80.50}&{83.15}&{81.80}\\
\multirow{2}{*}{}&{(72.13)}&{(67.46)}&{(69.71)}&{(80.92)}&{(76.78)}&{(78.79)}&{(83.59)}&{(80.34)}&{(81.92)}\\
\hline

\end{tabular}
\end{table*}

\section{Results and Analyses}
\subsection{Comparisons with the Baselines}
The main results of the baseline models and our model in three discontinuous NER datasets are shown in Table \ref{tab:table3}. We can observe that our model has the best results (F1 values) on the three datasets, which is due to the fact that it not only captures the relationships between words, but also pays attention to the relationships between words and tags and the relationships between tags and tags. 
In addition, we expand two new tags, which can complement the prediction of neighbor words and head-tail words. 
As a result, the performance on CADEC, ShARe13 and ShARe14 is improved by 0.83\%, 0.05\% and 0.66\% compared to the previous state-of-the-art ones. 

\subsection{Ablation Studies}
In addition, we also conduct corresponding ablation experiments, in order to verify the effectiveness of the modules in our proposed model and understand their impacts. The experimental results are shown in Table \ref{tab:table4}. First of all, if the TREM module is deleted, the F1s of our model on three datasets decrease by {1.24\%, 0.59\% and 1.03\%} respectively. This shows that the TREM is effective, that is, the relationships between words and tags are conducive to the prediction of tags.


Then, we also investigate the effect of the iteration number in TREM. As can be seen from the table, no matter reducing the number of iterations (e.g. rounds = 2) or increasing the number of iterations (e.g. rounds = 4), the performance of our model on all three datasets decrease. 
Specifically, when we use 2 rounds, the F1 decreases by {0.79\%, 1.19\% and 0.78\%} respectively.
Similarly, when 4 rounds are used, the F1 also decreases by {1.81\%, 1.16\% and 0.48\%} respectively.


After that, we conduct experiments to compare the effect of using the tags in the SOTA model \cite{ref21} (NNW+THW) and the ones that we propose in this paper (PNW+HTW).
As shown in Table \ref{tab:table4}, the NNW+THW and PNW+HTW tagging strategies achieve almost the same F1s, which is explainable because they entail the same neighbor-word and head-tail relationships but implement these relationships in different directions.
The two new tags that we have added are kinds of effective complements to the previous tags.
Therefore, the model that combines both of them achieves the best performance on all three datasets.

\begin{figure*}[!t]
\centering
\includegraphics[scale=0.5]{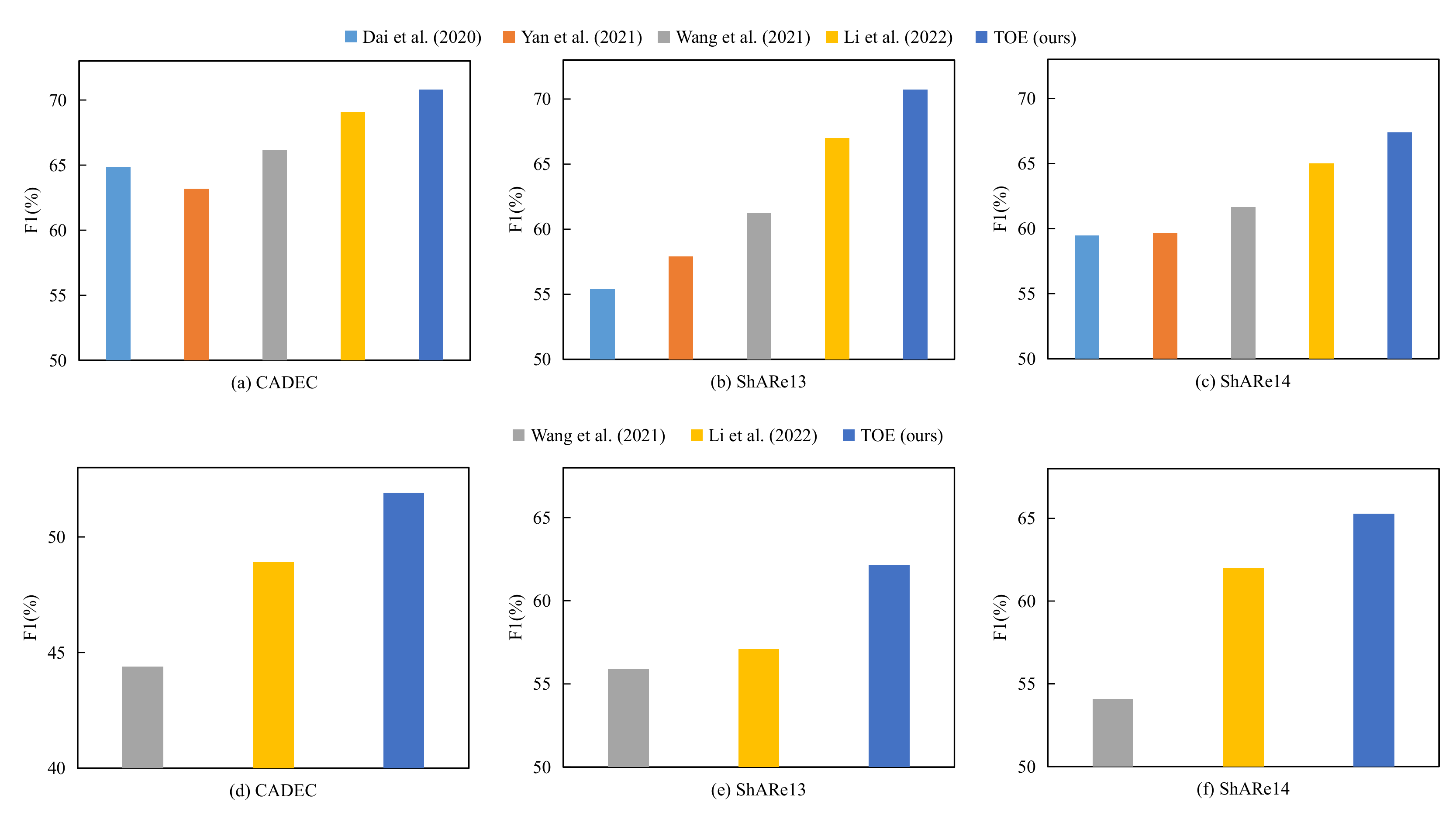}
\caption{Results of recognizing discontinuous entities, where (a) - (c) are the F1 values using the sentences containing at least one discontinuous entity, and (d) - (f) are the F1 values considering only discontinuous entities.} 
\label{fig4}
\end{figure*}

\begin{table}[t]
\centering
\caption{Examples of different overlapped types.
} 
\includegraphics[scale=0.36]{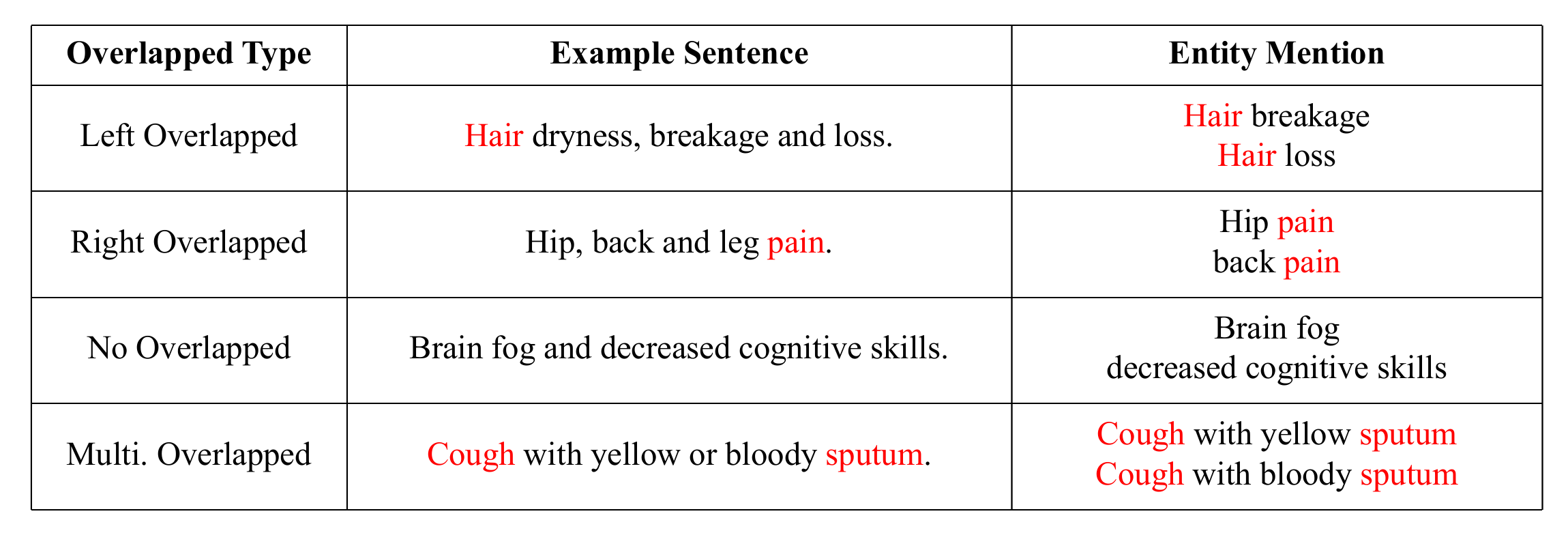}
\label{fig5}
\end{table}

Next, we investigate the effect of using different decoding methods. As shown in Table \ref{tab:table4}, the ``T(NNW, PNW) L(THW, HTW)'' decoding method performs the best.
We assume that the reason for this observation is that NNW and PNW relationships occur much more than THW and HTW relationships.
Thus, the model is apt to predict more NNW and PNW relationships but fewer THW and HTW relationships.
Therefore, it is necessary to tighten the establishing condition of neighbor word relationships by using the logic ``AND'' between NNW and PNW, but loose the establishing condition of head-tail word relationships by using the logic ``OR'' between THW and HTW.
As seen, the ``L(NNW, PNW) T(THW, HTW)'' decoding method, which is the opposite of the ``T(NNW, PNW) L(THW, HTW)'' decoding method, performs the worst in all three datasets, further verifying our assumption.

Finally, we also investigate the effect of the tag mapping number. We test 4 values for the tag mapping number $M_{num}$, namely 2, 4, 6, 8. We find that when the tag mapping number is set to 4, which is the same as the number of tags, our model achieves the best performance.
Such an observation is consistent with our intuition that the tag mapping number should be equal to the tag number.
This may be because each tag mapping represents a tag, more or fewer tag mappings may bring troubles to the model for aligning tag representations with tags.


\begin{table}[!t]
\caption{Experimental results of recognizing the entities with different overlapped types.
The bold number denotes the highest value for each type and dataset.
\label{tab:table5}}
\centering
\begin{tabular}{c |c| c c c}
\hline
\multicolumn{2}{c|}{{\bf{Model}}}&{\bf{CADEC}}&{\bf{ShARe13}}&{\bf{SHAER14}}\\
\hline
\multirow{4}{*}{Wang et al.(2021) \cite{ref20}}&{\bf{No}}&7.69 &	48.26& 	40.32\\
\multirow{4}{*}{}&{\bf{Left}}&42.86 &	64.07& 	66.15 
 \\
\multirow{4}{*}{}&{\bf{Right}}&{\bf{62.22}} &	13.04& 	45.80 
 \\
\multirow{4}{*}{}&{\bf{Multi.}}&0.00 &	29.63 &	0.00\\ 
 \hline
\multirow{4}{*}{Li et al.(2022) \cite{ref21}}&{\bf{No}}&32.79 &	48.48 &	50.38 
\\
\multirow{4}{*}{}&{\bf{Left}}&46.51 &	69.60 &	62.40 
 
 \\
\multirow{4}{*}{}&{\bf{Right}}&51.43 &	{\bf{58.89}} &	68.77 
 
 \\
\multirow{4}{*}{}&{\bf{Multi.}}&{\bf{17.98}}& 	33.33 &	0.00 
\\ 
\hline
\multirow{4}{*}{TOE(ours)}&{\bf{No}}&{\bf{38.87}} &	{\bf{50.21}}&{\bf{53.29}} 
\\
\multirow{4}{*}{}&{\bf{Left}}&{\bf{48.48}} &{\bf{69.71}} &	{\bf{66.32}} 
 
 \\
\multirow{4}{*}{}&{\bf{Right}}&56.79 &	49.59 &	{\bf{72.31}}
 
 \\
\multirow{4}{*}{}&{\bf{Multi.}}&0.00 &{\bf{36.36}} &	0.00 
\\ 
\hline
\end{tabular}
\end{table}
\subsection{Effectiveness on Recognizing Discontinuous Entities}

Figures \ref{fig4}(a) - (c) show the results using sentences containing at least one discontinuous entity. First, our model TOE achieves better results than other grid-tagging models such as Wang et al. (2021) \cite{ref20} and Li et al. (2022) \cite{ref21}.
This demonstrates the superiority of our model, where both the word and tag relationships are leveraged.
In addition, our model also performs better than other kinds of baselines such as the transition-based system \cite{ref54} and end-to-end generative model \cite{ref17}.
Moreover, as shown in Figures \ref{fig4}(d) - (f), when only comparing the performance of recognizing discontinuous entities, our model still obtains the best results on all datasets. In conclusion, our model has the advantage of identifying discontinuous entities, which contributes to overall performance improvement.


\begin{table*}[!t]
\caption{Error analysis on different entity types. Head-tail relationship corresponds to the THW and HTW tags, and the neighbor words relationship corresponds to the NNW and PNW tags.
All results come from sentences containing at least one corresponding entity type.
\label{tab:table6}}
\centering
\begin{tabular}{c |c |c |c |c |c}
\hline
{\bf{Entity Type}}&\multicolumn{2}{c|}{{\bf{Error Type}}}&{\bf{CADEC(\%)}}&{\bf{ShARe13(\%)}}&{\bf{ShARe14(\%)}}\\
\hline\hline
\multirow{5}{*}{All}&\multirow{2}{*}{FP}&Head-tail relationship correct, neighbor words relationship incorrect&4.05 &	0.50& 	1.01\\ 
\cline{3-6}
\multirow{5}{*}{}&\multirow{2}{*}{}&Head-tail relationship incorrect&40.28 &	40.33& 	52.29 \\
\cline{2-6}
\multirow{5}{*}{}&\multirow{2}{*}{FN}&Head-tail relationship correct, neighbor words relationship incorrect&2.43&0.50&0.91 \\
\cline{3-6}
\multirow{5}{*}{}&\multirow{2}{*}{}&Head-tail relationship incorrect	&		53.24& 	58.67& 	45.79\\
\cline{2-6}
\multirow{5}{*}{}&\multicolumn{2}{c|}{Total}&100&	100&	100\\
\hline\hline
\multirow{5}{*}{Discontinuous Entity}&\multirow{2}{*}{FP}&Head-tail relationship correct, neighbor words relationship incorrect&9.74 &	5.06& 	6.87 
\\ 
\cline{3-6}
\multirow{5}{*}{}&\multirow{2}{*}{}&Head-tail relationship incorrect&15.48 &	17.05 &	35.23 
 \\
\cline{2-6}
\multirow{5}{*}{}&\multirow{2}{*}{FN}&Head-tail relationship correct, neighbor words relationship incorrect&5.73 &	3.94 &	4.65 
\\
\cline{3-6}
\multirow{5}{*}{}&\multirow{2}{*}{}&Head-tail relationship incorrect	&69.05 &	73.95 &	53.25 
\\
\cline{2-6}
\multirow{5}{*}{}&\multicolumn{2}{c|}{Total}&100&	100&	100\\
\hline\hline
\multirow{5}{*}{Flat Entity}&\multirow{2}{*}{FP}&Head-tail relationship correct, neighbor words relationship incorrect&0.00& 	0.00 &	0.00 
\\ 
\cline{3-6}
\multirow{5}{*}{}&\multirow{2}{*}{}&Head-tail relationship incorrect&50.14 &	39.96 &	58.92 
 \\
\cline{2-6}
\multirow{5}{*}{}&\multirow{2}{*}{FN}&Head-tail relationship correct, neighbor words relationship incorrect&0.00 &	0.00 &	0.00 
\\
\cline{3-6}
\multirow{5}{*}{}&\multirow{2}{*}{}&Head-tail relationship incorrect	&49.86 &	60.04 &	41.08 
\\
\cline{2-6}
\multirow{5}{*}{}&\multicolumn{2}{c|}{Total}&100&	100&	100\\
\hline\hline
\multirow{5}{*}{Overlapped Entity}&\multirow{2}{*}{FP}&Head-tail relationship correct, neighbor words relationship incorrect&12.53 &	4.70 &	3.47 
\\ 
\cline{3-6}
\multirow{5}{*}{}&\multirow{2}{*}{}&Head-tail relationship incorrect&28.46 &	17.10 &	34.74 
\\
\cline{2-6}
\multirow{5}{*}{}&\multirow{2}{*}{FN}&Head-tail relationship correct, neighbor words relationship incorrect&1.76 &	4.27 &	2.65 
\\
\cline{3-6}
\multirow{5}{*}{}&\multirow{2}{*}{}&Head-tail relationship incorrect	&57.25& 	73.93 &	59.14 
\\
\cline{2-6}
\multirow{5}{*}{}&\multicolumn{2}{c|}{Total}&100&	100&	100\\
\hline
\end{tabular}
\end{table*}

\begin{table}[!t]
\caption{Efficiency Analysis. Sent/S is the number of sentences that can be processed per second.\label{tab:table7}}
\centering
\begin{tabular}{c|c c}
\hline
\multirow{2}{*}{{\bf{Model}}}&{\bf{Training}}&{\bf{Inference}}\\
\multirow{2}{*}{}&{\bf{(sent/s)}}&{\bf{(sent/s)}}\\
\hline
Dai et al.(2020) \cite{ref54}&		24.7 &	66.5 \\
Yan et al.(2021) \cite{ref17}&		63.6 &	19.2 \\
Wang et al.(2021) \cite{ref20}&	39.3 &	109.7 \\
Li et al.(2022) \cite{ref21}	&	116.1 &	365.7 \\
\hline
TOE(ours)&	78.5 &	195.1 \\
\hline
\end{tabular}
\end{table}
\subsection{Performance Analysis of Recognizing the Entities with Different Overlapped Types}
As mentioned in the previous section, discontinuous entities and overlapped entities basically exist at the same time in the three datasets. In order to analyze the ability of our model to extract various overlapping entities, we divide them into four categories according to previous work \cite{ref54, ref20}, i.e., ``no overlapped'', ``left overlapped'', ``right overlapped'' and ``multiple overlapped''. 
Table \ref{fig5} shows an example for each overlapped category.
As shown in Table \ref{tab:table5}, most of the results of our model are optimal compared with the baseline models, with regards to different overlapped types. 
This shows that our model is more competitive and more adaptive in extracting various overlapping entities.
In addition, in some cases, the F1s of our model and the baseline models are 0. This may be because the number of such entities is significantly smaller in the datasets, compared to the ones of other overlapping types.
For instance, the number of multiple overlapped entities in the training set of ShARe14 is 20, leading to under-fitting for model training. 

\subsection{Error Analysis}
In order to understand the disadvantages and advantages of our model, we perform an error analysis and show the results in this section.
Errors can be divided into false-positive (FP) and false-negative (FN). 
Furthermore, FP errors and FN errors can be further divided into two types: ``head-tail relationship incorrect'' and ``head-tail relationship correct but neighbor words relationship incorrect''. 
As shown in Table \ref{tab:table6}, we analyze the errors of our model on three discontinuous datasets from four directions: ``all entities'', ``flat entities'', ``overlapped entities'' and ``discontinuous entities''. 
We can see that regardless of the entity type, the FN and FP errors for ``head-tail relationship incorrect'' account for high proportions.
Especially in recognizing flat entities, 100\% errors come from incorrect head-tail relationship recognition. 
This may be because the head-tail relationship is a more difficult relationship to recognize compared with the neighbor relationship, since entity boundary is hard to identify, which is a well-known issue in previous work \cite{ref60, ref18}.
In addition, we can also observe that in most cases, the FP numbers of ``head-tail relationship incorrect'' are smaller than the FN numbers, which shows that the recognition coverage of gold-standard entities is still a challenge for our model.



\subsection{Efficiency Analysis}
Table \ref{tab:table7} shows the training speed and prediction speed of our model. Compared with the model of Li et al. (2022) \cite{ref21}, our model has a slower training speed and prediction speed. This is mainly because we inject the tag embedding into our model and the computation in the TREM can be performed iteratively, resulting in more parameters and calculations. 
Although the performance improvement of our model leads to a decrease in efficiency compared with the model proposed by Li et al. (2022) \cite{ref21}, our model still has certain advantages in the training and prediction speeds compared with other baseline models \cite{ref54, ref17, ref20}, which demonstrates that the method based on grid tagging is more efficient than other kinds of methods.

\section{Conclusion}
This paper extends a SOTA grid-tagging model for discontinuous named entity recognition
with tag-oriented enhancement. 
Our enhanced model has two strengths:
(1) It not only pays attention to the relationships between words, but also the relationships between words and tags.
(2) It leverages a more fined-grain tagging system to strengthen the prediction of the relationships between words and tags. 
The experimental results show that the performance of our model on all three benchmark datasets are the best, and the ablation experiments demonstrate the effectiveness of the two enhancements that we propose in the model. Further experimental analyses show that our proposed model can better identify discontinuous entities. 
Although the enhancements bring a certain efficiency loss, our model is still faster than most baselines in training and prediction.
In the future, we will apply our model in more complex information extraction tasks such as nested entity relationship extraction and structured sentiment analysis.

\section*{Acknowledgments}
This work is supported by the National Natural Science Foundation of China (No. 62176187), the National Key Research and Development Program of China (No. 2017YFC1200500), the Research Foundation of Ministry of Education of China (No. 18JZD015), the Youth Fund for Humanities and Social Science Research of Ministry of Education of China (No. 22YJCZH064), the General Project of Natural Science Foundation of Hubei Province (No.2021CFB385).
This work is also the research result of the independent scientific research project (humanities and social sciences) of Wuhan University, supported by the Fundamental Research Funds for the Central Universities.
L Zhao would like to thank the support from Center for Artificial Intelligence (C4AI-USP), the Sao Paulo Research Foundation (FAPESP grant \#2019/07665-4), the IBM Corporation, and China Branch of BRICS Institute of Future Networks.

\bibliography{ref}
\bibliographystyle{IEEEtran}

\end{CJK}
\end{document}